# Augmenting Inertial Motion Capture with SLAM Using EKF and SRUKF Data Fusion Algorithms

Mohammad Mahdi Azarbeik[1], Hamidreza Razavi, Kaveh Merat, Hassan Salarieh

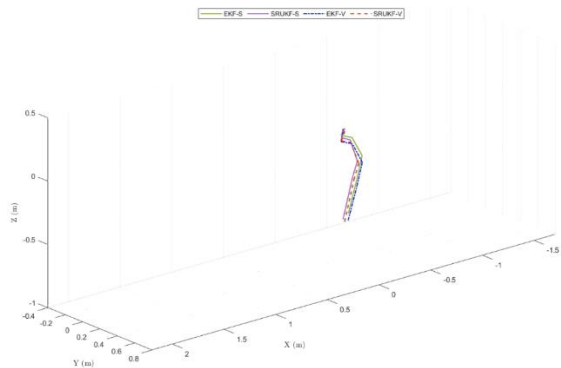

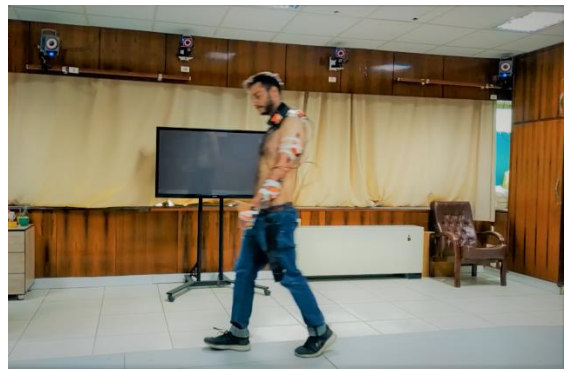

*Abstract*— **Inertial motion capture systems widely use low-cost IMUs to obtain the orientation of human body segments, but these sensors alone are unable to estimate link positions. Therefore, this research used a SLAM method in conjunction with inertial data fusion to estimate link positions. SLAM is a method that tracks a target in a reconstructed map of the environment using a camera. This paper proposes quaternion-based extended and square-root unscented Kalman filters (EKF & SRUKF) algorithms for pose estimation. The Kalman filters use measurements based on SLAM position data, multi-link biomechanical constraints, and vertical referencing to correct errors. In addition to the sensor biases, the fusion algorithm is capable of estimating link geometries, allowing the imposing of biomechanical constraints without *a priori* knowledge of sensor positions. An optical tracking system is used as a reference of ground-truth to experimentally evaluate the performance of the proposed algorithm in various scenarios of human arm movements. The proposed algorithms achieve up to 5.87 (*cm*) and 1.1 (*deg*) accuracy in position and attitude estimation. Compared to the EKF, the SRUKF algorithm presents a smoother and higher convergence rate but is 2.4 times more computationally demanding. After convergence, the SRUKF is up to 17% less and 36% more accurate than the EKF in position and attitude estimation, respectively. Using an absolute position measurement method instead of SLAM produced 80% and 40%, in the case of EKF, and 60% and 6%, in the case of SRUKF, less error in position and attitude estimation, respectively.**

## I. Introduction

SEVERAL techniques have been developed to monitor human body motions. These techniques differ in biomedical applications, virtual and augmented reality, human-machine interaction, and the film industry. Inertial motion capture using inertial measurement units (IMUs) has become more attractive than traditional marker-based optical techniques because of its low cost, small sensor size, and ability to be used in non-laboratory environments [1]. Although optical motion tracking systems are accurate, they are mostly limited to laboratory environments and interior applications. Additionally, visual sensors suffer from occlusion when targets are obstructed by other joints, limbs, or objects [2]. Another motion capture method relies on Microsoft Kinect [3] which, like other optical systems, suffers from occlusion and has a low sampling rate [4].

In [5], the data from multiple Kinect sensors are merged to track human movements because the subject must always face Kinect to overcome the occlusion problem.

Body link orientations are estimable using data from IMUs attached to body links [6]. However, IMUs suffer from uncalibrated sensor [7] and drift [8] problems, especially in dynamic motions. Furthermore, human body localization is not feasible through IMU data alone. Various studies have been conducted to determine the orientation of IMU sensors. Complimentary filters with low computational costs and increased accuracy in attitude estimation are proposed in [9] and [10]. In [11], the authors proposed a method for estimating arm attitude by fusing the measurements of an IMU and a Kinect sensor using an unscented Kalman filter. Due to the limited visual range of the Kinect sensor, the authors assessed the accuracy of the proposed algorithm without the assistance of Kinect in [12]. Inertial motion capture algorithms capable of estimating sensor parameters and link geometry in real-time based on extended and unscented Kalman filtering were introduced and compared in [13].

---
[1] M. M. Azarbeik is with the Department of Mechanical Engineering, Sharif University of Technology, Tehran, Iran. (e-mail: mm.azarbeik@gmail.com, mohammamahdi.azarbeik@mech.sharif.edu)

Intensive research on simultaneous localization and mapping (SLAM) and visual odometry (VO) systems using cameras alone, or combined with inertial sensors, have produced methods to estimate the position and orientation of a particular target [14]. SLAM methods construct or update a map of an unknown environment while simultaneously tracking the position of an agent. A system capable of performing visual, visual-inertial, and multi-map SLAM with monocular, stereo, and RGB-D cameras, using pin-hole and fisheye lens models, "ORB-SLAM3", is proposed in [14]. In [15], the semidirect VO "SVO", which is significantly faster than many of the state-of-the-art VO algorithms, is proposed. While achieving highly competitive accuracies, SVO is considerably less accurate than ORB-SLAM3. As stated, IMU data alone is insufficient for body links' position estimation. Therefore, this research uses a SLAM method alongside these sensors to estimate the position of links.

In this study, an inertial motion capture algorithm combined with a SLAM method (ORB-SLAM3 [14]), based on extended and square-root unscented Kalman filtering, capable of estimating the positions and attitudes of multiple body links in unknown, outdoor, and non-laboratory environments, with appropriate accuracy, is proposed. Due to the well-defined error models to minimize the effects of measurement noise on the corresponding error dynamics, simple sensor fusion algorithms such as EKF are able to handle the system nonlinearities. However, the performance of the Kalman filter in parameter estimation correction is evaluated using extended and unscented Kalman filters due to different convergence rates. As the conventional unscented Kalman filter suffers from numerical stability issues, a square-root covariance propagation and update algorithm is used to mitigate this issue. The computational cost of the unscented Kalman filter increases substantially as the number of the estimated states increases. Gyroscopes and accelerometers biases, which for low-cost IMUs tend to change with temperature and passage of time, are estimated alongside the position, velocity, attitude, and geometry of body links. The ability to estimate body links geometry allows the imposition of biomechanical constraints without *a priori* knowledge of the sensors and the camera positioning [16]. Additionally, stationary accelerometer data are used as a complementary measurement to the filter for attitude correction. Combining multi-link inertial motion capture with the SLAM method for human body localization and pose estimation making the motion capture system practical in any environment is the main innovation of the present study. The VICON optical motion capture system is used both as a verification ground-truth and an alternative position measurement for the introduced algorithms.

## II. FUNDAMENTALS

This section describes the system overview, various reference frames, dynamic equations, and sensor output models used throughout the paper which are based on [13].

The multi-body system the motion of which is to be captured consists of $N$ links, each being connected to at least one other link via a ball and socket joint. The vectors connecting the sensor modules of $i$-th and $j$-th links to their connecting joint, $J_{ij}$, will be denoted by segment $l_{ij}^{b_i} \in \mathbb{R}^3$ and $l_{ji}^{b_j} \in \mathbb{R}^3$ (Fig. 1) [13]. The camera is assumed to be installed on the $0$-th module and the lever arm connecting the $0$-th module IMU to the camera is denoted by $l_c^{b_0} \in \mathbb{R}^3$. The recurring symbols used throughout the paper are tabulated in Table I.

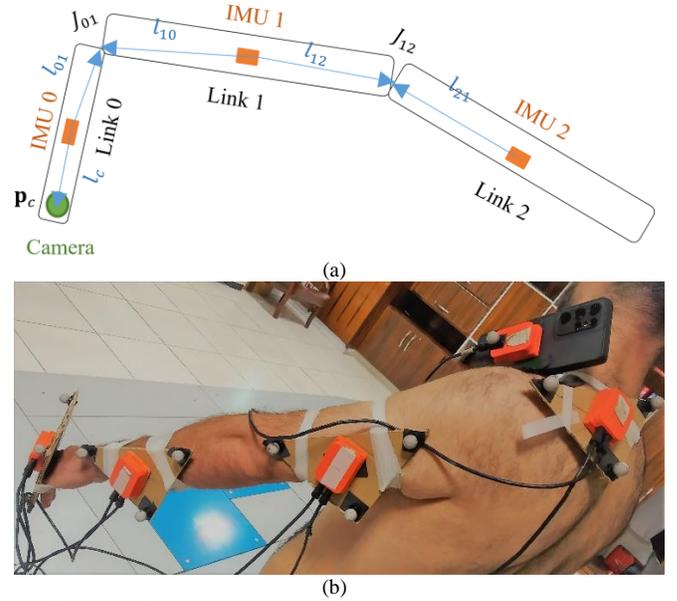

Fig. 1. (a) Schematic of sensors, links, and markers placement. (b) Experimental setup.

TABLE I
NOMENCLATURE

| Symbol | Description |
|---|---|
| $J_{ij}$ | The joint between the $i$-th and $j$-th links |
| $l_{ij}^{b_i}$ | The vector connecting the $i$-th module to $J_{ij}$ |
| $l_c^{b_0}$ | Lever arm connecting the $0$-th module IMU to the camera |
| $n$ | NED (North-East-Down) frame |
| $b_k$ | $k$-th module sensor body frame |
| $p_k$ | $k$-th module's estimation of the $n$-frame |
| $\cdot_c$ | Variable is associated with the camera |
| $\tilde{\cdot}$ | Measured value |
| $\hat{\cdot}$ | Estimated value |

In this article, the direction cosine matrix (DCM), quaternion, and rotation vector described in [17] are used to parameterize rotations. The symbol $\otimes$ refers to the quaternion product defined as,

$$\mathbf{q}_1 \otimes \mathbf{q}_2 = \begin{bmatrix} q_1 q_2 - \mathbf{q}_{1v} \cdot \mathbf{q}_{2v} \\ q_1 \mathbf{q}_{2v} + q_2 \mathbf{q}_{1v} + \mathbf{q}_{1v} \times \mathbf{q}_{2v} \end{bmatrix} \quad (1)$$

$$\mathbf{q}^T = \begin{bmatrix} q & \mathbf{q}_v^T \end{bmatrix} \quad (2)$$

where $q_1, q_2, q$ and $\mathbf{q}_{1v}, \mathbf{q}_{2v}, \mathbf{q}_v \in \mathbb{R}^3$ represent the scalar and vector parts of unit quaternions $\mathbf{q}_1$, $\mathbf{q}_2$, and $\mathbf{q}$, respectively.

For an arbitrary vector $\mathbf{z}$ the symbols $\hat{\mathbf{z}}$ and $\tilde{\mathbf{z}}$ indicate the estimated and measured values of $\mathbf{z}$, respectively, and $\delta\mathbf{z}$ denotes the estimation error vector.

$$\mathbf{z} = \delta\mathbf{z} + \hat{\mathbf{z}} \tag{3}$$

The quaternion form of vector $\mathbf{d} \in \mathbb{R}^3$ is represented by,

$$\overline{\mathbf{d}}^T = \begin{bmatrix} 0 & \mathbf{d}^T \end{bmatrix} \tag{4}$$

The $n$-frame denotes the NED (North-east-down) frame, the $b_k$-frame refers to the $k$-th module sensor body frame, and the $p_k$-frame is the $k$-th module's estimation of the $n$-frame which satisfies,

$$\hat{\mathbf{R}}_{b_k}^n = \mathbf{R}_{b_k}^{p_k} \tag{5a}$$

$$\hat{\mathbf{q}}_{b_k}^n = \mathbf{q}_{b_k}^{p_k} \tag{5b}$$

where $\hat{\mathbf{q}}_{b_k}^n$ and $\mathbf{q}_{b_k}^{p_k}$ are the quaternion equivalents of DCMs $\hat{\mathbf{R}}_{b_k}^n$ and $\mathbf{R}_{b_k}^{p_k}$, respectively.

In this study, the effects of the earth's rotational velocity will be neglected due to the relatively larger measurement noise and resolution. Consequently, the dynamic model equations of the $k$-th module are obtained by simplifying the equations of the inertial navigation system (INS) [17],

$$\dot{\mathbf{p}}_k^n = \mathbf{v}_k^n \tag{6a}$$

$$\dot{\mathbf{v}}_k^n = \mathbf{R}_{b_k}^n \mathbf{f}_k^{b_k} + \mathbf{g}^n \tag{6b}$$

$$\dot{\mathbf{q}}_{b_k}^n = \frac{1}{2} \mathbf{q}_{b_k}^n \otimes \overline{\omega}_{b_k}^{b_k} \tag{6c}$$

where the vectors $\mathbf{p}_k^n, \mathbf{v}_k^n \in \mathbb{R}^3$ denote the position and velocity of the $k$-th module in the $n$-frame, respectively. The gravity vector in the $n$-frame is denoted as $\mathbf{g}^n \in \mathbb{R}^3$. The vectors $\mathbf{f}_k^{b_k}, \overline{\omega}_{b_k}^{b_k} \in \mathbb{R}^3$ refer to the specific force and rotational velocity of the $k$-th module expressed in the $b_k$-frame, respectively. The unit quaternion $\mathbf{q}_{b_k}^n \in \mathbb{R}^4$ and rotation matrix $\mathbf{R}_{b_k}^n \in \mathbb{R}^{3\times3}$ parametrize the $b_k$-frame attitude in relation to the $n$-frame.

Due to sensor bias, scaling, non-orthogonality, and measurement noise, the accelerometer and gyroscope outputs differ from the actual values of specific force and angular velocity. As shown in [13], the estimation accuracy in inertial motion capture scenarios is not necessarily enhanced by adaptive estimation of accelerometer and gyroscope scaling and non-orthogonality parameters due to their poor observability. Therefore, the IMUs outputs are modeled by,

$$\mathbf{f}_k^{b_k} = \tilde{\mathbf{f}}_k^{b_k} - \mathbf{b}_{a,k} - \eta_{a,k} \tag{7a}$$

$$\omega_{b_k}^{b_k} = \tilde{\omega}_{b_k}^{b_k} - \mathbf{b}_{g,k} - \eta_{g,k} \tag{7b}$$

where the indexes $(.)_{a,k}$ and $(.)_{g,k}$ correspond respectively to the accelerometer and the gyroscope of the $k$-th module. The vectors $\mathbf{b} \in \mathbb{R}^3$ and $\eta \in \mathbb{R}^3$ represent the bias and white Gaussian measurement noise vectors, respectively. The standard deviations (SD) of $\eta_{a,k}$ and $\eta_{g,k}$ are $\sigma_{a,k} > 0$ and $\sigma_{g,k} > 0$, respectively. According to (7),

$$\hat{\mathbf{f}}_k^{b_k} \approx \tilde{\mathbf{f}}_k^{b_k} - \hat{\mathbf{b}}_{a,k} \tag{8a}$$

$$\hat{\omega}_{b_k}^{b_k} \approx \tilde{\omega}_{b_k}^{b_k} - \hat{\mathbf{b}}_{g,k} \tag{8b}$$

### III. ERROR MODELS

Considering (3), (7), and (8) the sensor output errors are modeled as,

$$\delta\mathbf{f}_k^{b_k} = \mathbf{f}_k^{b_k} - \hat{\mathbf{f}}_k^{b_k} = -\delta\mathbf{b}_{a,k} - \eta_{a,k} \tag{9a}$$

$$\delta\omega_{b_k}^{b_k} = \omega_{b_k}^{b_k} - \hat{\omega}_{b_k}^{b_k} = -\delta\mathbf{b}_{g,k} - \eta_{g,k} \tag{9b}$$

The attitude quaternion estimation error, unlike the other system parameters which follow (3), is modeled as,

$$\mathbf{q}_{b_k}^n = \mathbf{q}_{p_k}^n \otimes \mathbf{q}_{b_k}^{p_k} \tag{10}$$

where $\mathbf{q}_{p_k}^n$ represents the $n$-frame attitude estimation error and is used to rotate its axes to obtain the $p_k$-frame axes. The rotation vector equivalent of $\mathbf{q}_{p_k}^n$ is indicated by $\phi_k$.

The present inertial motion capture algorithm is implemented using both extended and unscented Kalman filters. Since the unscented implementation calculates sigma points, it only requires the parameter estimation error models, system dynamics, and measurement equations. However, the extended implementation requires the estimation error dynamics and error-based measurement residuals [18]. To obtain the estimation error dynamics required for the extended Kalman filter, $\phi_k$ is assumed to be small (i.e. $\|\phi_k\| << 1$). Therefore, $\mathbf{R}_{p_k}^n$, the DCM representing the rotation between the $n$-frame and $p_k$-frame, and its quaternion equivalent, $\mathbf{q}_{p_k}^n$, satisfy these equations [17].

$$\mathbf{R}_{p_k}^n \approx \mathbf{I}_{3\times3} + [\phi_k \times] \tag{11a}$$

$$(\mathbf{q}_{p_k}^n)^T \approx [1 \quad \tfrac{1}{2}\phi_k^T] \tag{11b}$$

Subsequently, differentiating (10) with respect to time, while considering (6c) and (11b), we have

$$\dot{\mathbf{q}}_{p_k}^n = \frac{1}{2} \mathbf{q}_{p_k}^n \otimes \delta\overline{\omega}_{b_k}^{p_k} \tag{12}$$

Considering (9b), (11b), (12), small angle rotation vector dynamics, and omitting higher order terms results in [17],

$$\dot{\phi}_k \approx \delta\omega_{b_k}^{p_k} \tag{13}$$

Taking into account (3), differentiating the velocity estimation with respect to time and substituting by (6b), (7a), (9a), and (11a) leads to,

$$\delta\dot{\mathbf{v}}_k^n = \delta\mathbf{f}_k^{p_k} - \hat{\mathbf{f}}_k^{p_k} \times \phi_k \tag{14}$$

Taking into account (3), differentiating the position estimation error with respect to time and substitution of (6a) results in:

$$\delta\dot{\mathbf{p}}_k^n = \delta\mathbf{v}_k^n \tag{15}$$

### IV. SENSOR FUSION

As a result of measurement noise, incorrect parameter initialization, and numerical integration, position, velocity, and attitude estimation of each link using the differential equations of (6) are prone to error. In both extended and unscented

Kalman filters, measurements based on biomechanical constraints present in a multi-link system, SLAM output using the camera, and vertical referencing using accelerometer measurements are used to correct these errors. The position and velocity of each joint are obtainable using the position, velocity, and attitude of the adjacent link module. Considering (Fig. 1),

$$\mathbf{p}_{J_{ij,i}}^n = \mathbf{p}_i^n + \mathbf{R}_{b_i}^n \mathbf{l}_{ij}^{b_i} \tag{16}$$

where $\mathbf{p}_{J_{ij,i}}^n$ is the position of the joint $J_{ij}$ obtained from the $i$-th link parameters. Ideally, $\mathbf{p}_{J_{ij,i}}^n$ and $\mathbf{p}_{J_{ij,j}}^n$ should be equivalent, however, the body joints deform slightly during motions. For the purpose of imposing the equivalence constraint between $\mathbf{p}_{J_{ij,i}}^n$ and $\mathbf{p}_{J_{ij,j}}^n$, the following pseudo-measurement vector is defined considering (16) to model the joints' deformations,

$$\tilde{\mathbf{y}}_{P_{ij}} = \mathbf{p}_j^n - \mathbf{p}_i^n + \mathbf{R}_{b_j}^n \mathbf{l}_{ji}^{b_j} - \mathbf{R}_{b_i}^n \mathbf{l}_{ij}^{b_i} + \eta_{p,J_{ij}} \tag{17}$$

where $\eta_{p,J_{ij}} \in \mathbb{R}^3$ is a random Gaussian noise vector, with standard deviation $\sigma_{p,J_{ij}} > 0$. Considering (5a) and (17),

$$\hat{\mathbf{y}}_{P_{ij}} = \hat{\mathbf{p}}_j^n - \hat{\mathbf{p}}_i^n + \mathbf{R}_{b_j}^{p_j} \hat{\mathbf{l}}_{ji}^{b_j} - \mathbf{R}_{b_i}^{p_i} \hat{\mathbf{l}}_{ij}^{b_i} \tag{18}$$

The joint position residual vector, $\delta\mathbf{y}_{P_{ij}} \in \mathbb{R}^3$, is defined as,

$$\delta\mathbf{y}_{P_{ij}} = \tilde{\mathbf{y}}_{P_{ij}} - \hat{\mathbf{y}}_{P_{ij}} \tag{19}$$

substituting from (11a), (17), and (18) into (19) and disregarding higher order terms,

$$\delta\mathbf{y}_{P_{ij}} \approx \delta\mathbf{p}_j^n - \delta\mathbf{p}_i^n + \eta_{p,J_{ij}} + \mathbf{R}_{b_j}^{p_j} \delta\mathbf{l}_{ji}^{b_j} - \mathbf{R}_{b_i}^{p_i} \delta\mathbf{l}_{ij}^{b_i} - \hat{\mathbf{l}}_{ji}^{p_j} \times \phi_j + \hat{\mathbf{l}}_{ij}^{p_i} \times \phi_i \tag{20}$$

where $\delta\mathbf{l}_{ij}^{b_i} \in \mathbb{R}^3$ and $\delta\mathbf{l}_{ji}^{b_j} \in \mathbb{R}^3$ are vectors of segment estimation errors. According to the joint velocity obtainable from the adjacent link module,

$$\mathbf{v}_{J_{ij,i}}^n = \mathbf{v}_i^n + \mathbf{R}_{b_i}^n (\omega_{b_i}^{b_i} \times \mathbf{l}_{ij}^{b_i}) \tag{21}$$

While $\mathbf{v}_{J_{ij,i}}^n$ is the velocity of joint $J_{ij}$ obtained from the parameters of the $i$-th link. Similar to the joint positions, $\mathbf{v}_{J_{ij,i}}^n$ and $\mathbf{v}_{J_{ij,j}}^n$ are not exactly equivalent as a result of joint deformations during motions. To enforce the equivalency constraint between $\mathbf{v}_{J_{ij,i}}^n$ and $\mathbf{v}_{J_{ij,j}}^n$ the next pseudo-measurement vector is defined by considering (21) to model joint deformations,

$$\tilde{\mathbf{y}}_{v_{ij}} = \mathbf{v}_j^n - \mathbf{v}_i^n + \eta_{v,J_{ij}} + \mathbf{R}_{b_j}^n (\omega_{b_j}^{b_j} \times \mathbf{l}_{ji}^{b_j}) - \mathbf{R}_{b_i}^n (\omega_{b_i}^{b_i} \times \mathbf{l}_{ij}^{b_i}) \tag{22}$$

where $\eta_{v,J_{ij}} \in \mathbb{R}^3$ is a random Gaussian noise vector, with standard deviation $\sigma_{v,J_{ij}} > 0$. Considering (5a) and (22),

$$\hat{\mathbf{y}}_{v_{ij}} = \hat{\mathbf{v}}_j^n - \hat{\mathbf{v}}_i^n + \mathbf{R}_{b_j}^{p_j} (\hat{\omega}_{b_j}^{b_j} \times \hat{\mathbf{l}}_{ji}^{b_j}) - \mathbf{R}_{b_i}^{p_i} (\hat{\omega}_{b_i}^{b_i} \times \hat{\mathbf{l}}_{ij}^{b_i}) \tag{23}$$

The joint velocity residual vector, $\delta\mathbf{y}_{v_{ij}} \in \mathbb{R}^3$, is defined as,

$$\delta\mathbf{y}_{v_{ij}} = \tilde{\mathbf{y}}_{v_{ij}} - \hat{\mathbf{y}}_{v_{ij}} \tag{24}$$

Substituting (11a), (22), and (23) to (24) and ignoring higher order terms results in,

$$\delta\mathbf{y}_{v_{ij}} \approx \delta\mathbf{v}_j^n - \delta\mathbf{v}_i^n + \eta_{v,J_{ij}} \\ -(\mathbf{R}_{b_j}^{p_j}(\hat{\omega}_{b_j}^{b_j} \times \hat{\mathbf{l}}_{ji}^{b_j})) \times \phi_j + (\mathbf{R}_{b_i}^{p_i}(\hat{\omega}_{b_i}^{b_i} \times \hat{\mathbf{l}}_{ij}^{b_i})) \times \phi_i \\ + \mathbf{R}_{b_j}^{p_j}(\delta\omega_{b_j}^{b_j} \times \hat{\mathbf{l}}_{ji}^{b_j} + \hat{\omega}_{b_j}^{b_j} \times \delta\mathbf{l}_{ji}^{b_j}) \\ - \mathbf{R}_{b_i}^{p_i}(\delta\omega_{b_i}^{b_i} \times \hat{\mathbf{l}}_{ij}^{b_i} + \hat{\omega}_{b_i}^{b_i} \times \delta\mathbf{l}_{ij}^{b_i}) \tag{25}$$

It's noteworthy that $\eta_{p,J_{ij}}$ and $\eta_{v,J_{ij}}$, in (17) and (22), are included to model the joints' deformations.

The measurement and estimation of $\mathbf{p}_c$, the position of the camera (Fig. 1), which can be obtained by the SLAM and the $k$-th module data, should be equivalent. To impose the equivalence constraint between these two, since the SLAM provides the direct position,

$$\tilde{\mathbf{y}}_{p_c} = \tilde{\mathbf{p}}_c^n = \mathbf{p}_0^n + \mathbf{R}_{b_0}^n \mathbf{l}_c^{b_0} + \eta_{p_c} \tag{26}$$

where $\tilde{\mathbf{p}}_c^n$ is the position obtained from the output of the SLAM and $\eta_{p_c} \in \mathbb{R}^3$ is a random Gaussian measurement noise vector, with standard deviation $\sigma_{p_c} > 0$. According to (5a) and (26),

$$\hat{\mathbf{y}}_{p_c} = \hat{\mathbf{p}}_c^n = \hat{\mathbf{p}}_0^n + \mathbf{R}_{b_0}^{p_0} \hat{\mathbf{l}}_c^{b_0} \tag{27}$$

The camera position residual vector, $\delta\mathbf{y}_{p_c} \in \mathbb{R}^3$, is defined as,

$$\delta\mathbf{y}_{p_c} = \tilde{\mathbf{y}}_{p_c} - \hat{\mathbf{y}}_{p_c} \tag{28}$$

substituting from (11a), (26), and (27) into (28) and disregarding higher order terms,

$$\delta\mathbf{y}_{p_c} \approx \delta\mathbf{p}_c^n - \delta\mathbf{p}_0^n - \eta_{p_c} - \mathbf{R}_{b_0}^{p_0} \delta\mathbf{l}_c^{b_0} + \hat{\mathbf{l}}_c^{p_0} \times \phi_0 \tag{29}$$

where $\delta\mathbf{l}_c^{b_0} \in \mathbb{R}^3$ is the camera lever arm estimation error vector. Notably, the SLAM output frequency is disordered, variable, and about one-third of the IMU's frequency. Thus, the algorithm uses measurement (29) each time the SLAM delivers data.

The estimated attitude for each module is corrected with the aid of the accelerometer measurements. Ideally, the output norms of static tri-axis accelerometers are close to the gravitational force. When the $k$-th link is steady, the gravity vector is obtainable as,

$$\mathbf{g}^n = -\mathbf{R}_{b_k}^n \mathbf{f}_k^{b_k} \tag{30}$$

Considering (5a), (30), and the estimated gravity vector during stationary periods,

$$\hat{\mathbf{g}}^n = -\mathbf{R}_{b_k}^{p_k} \hat{\mathbf{f}}_k^{b_k} \tag{31}$$

During stationary periods, the gravity vector and its estimate provide a vertical reference. Consequently, the gravity measurement residual, $\delta\mathbf{g}^n \in \mathbb{R}^3$, is defined as,

$$\delta\mathbf{g}^n = \mathbf{g}^n - \hat{\mathbf{g}}^n \tag{32}$$

Substituting (11a), (30), and (31) into (32) results in,

$$\delta\mathbf{g}^n \approx -\mathbf{g}^n \times \phi_k - \delta\mathbf{f}_k^n \tag{33}$$

The augmented state and estimation error vectors are denoted by **x** and **e**, respectively, and defined by,

$$\mathbf{x} = \begin{bmatrix} \mathbf{x}_1^T & \mathbf{x}_2^T & \cdots & \mathbf{x}_N^T & \mathbf{l}_c^{b_0 T} \end{bmatrix}^T, \\ \mathbf{e} = \begin{bmatrix} \mathbf{e}_1^T & \mathbf{e}_2^T & \cdots & \mathbf{e}_N^T & \delta\mathbf{l}_c^{b_0 T} \end{bmatrix}^T \tag{34}$$

Where $\mathbf{x}_k, \mathbf{e}_k$ $(k=1,\cdots,N)$ denote the augmented state and estimation error vectors of the $k$-th module and $N$ is the total number of the modules.

$$\begin{aligned}
\mathbf{x}_k &= \begin{bmatrix} {\mathbf{p}_k^n}^T & {\mathbf{v}_k^n}^T & {\mathbf{q}_{p_k}^n}^T & {\mathbf{b}_{a,k}}^T \cdots \\ \cdots {\mathbf{b}_{g,k}}^T & {\mathbf{l}_{k,k+1}^{b_k}}^T & {\mathbf{l}_{k,k-1}^{b_k}}^T \end{bmatrix}^T, \\
\mathbf{e}_k &= \begin{bmatrix} {\delta\mathbf{p}_k^n}^T & {\delta\mathbf{v}_k^n}^T & {\phi_k}^T & {\delta\mathbf{b}_{a,k}}^T \cdots \\ \cdots {\delta\mathbf{b}_{g,k}}^T & {\delta\mathbf{l}_{k,k+1}^{b_k}}^T & {\delta\mathbf{l}_{k,k-1}^{b_k}}^T \end{bmatrix}^T
\end{aligned} \quad (35)$$

Note that $\mathbf{x}_k$ and $\mathbf{e}_k$ respectively use unit quaternions and rotation vectors to parameterize the attitude.

An effective method for sensor fusion is the Kalman filter which has two main stages: time propagation of the states and error covariance by system dynamics, followed by states and error covariance correction by measurements. The performance of the Kalman filter in the correction of the parameter estimations obtained by the numerical integration of (6) is evaluated using extended and unscented schemes. The extended Kalman filter uses (13), (14), and (15) for attitude, velocity, and position error dynamics, respectively. The unscented Kalman filter is implemented using square-root covariance propagation and update for increased numerical stability similar to the method introduced in [19]. Due to the large number of the estimated states (e.g. 60 states, including 27 variables and 33 constants in the human arm motion tracking as discussed in section V), the SRUKF algorithm is implemented to mitigate the numerical instability issues encountered by the conventional UKF algorithm. The $\alpha$ and $\beta$ parameters that are responsible for sigma points distribution in the SRUKF algorithm [19] are considered to be 1 and 2, respectively. The related equations for the propagation and correction stages of each Kalman filter are tabulated in Table II.

TABLE II
KALMAN FILTER SCHEMES

| Filter | Propagation | Correction |
|---|---|---|
| EKF | (6), (13), (14), and (15) | (20), (25), (29), and (33) |
| SRUKF | (6) | (19), (24), (28), and (32) |

TABLE III
SENSOR NOISE CHARACTERISTICS

| Sensor | Random Walk | Bias Instability |
|---|---|---|
| Accelerometer | $60\ \mu g/\sqrt{hr}$ | $15\ \mu g$ |
| Gyroscope | $0.01\ ^\circ\!/s/\sqrt{hr}$ | $10\ ^\circ\!/h$ |

## V. EXPERIMENTAL RESULTS

In this section, the proposed algorithms' performance against a VICON optical motion capture system, which has a measurement error of less than 2mm [20], is experimentally evaluated by the human arm motions during 9 fast-paced gait and 3 jump motion scenarios on O-shaped and straight paths. The reference frame alignment between the inertial and optical systems is done using the method proposed in [21]. The subject traveled up to 200 meters in over 180 seconds during each test.

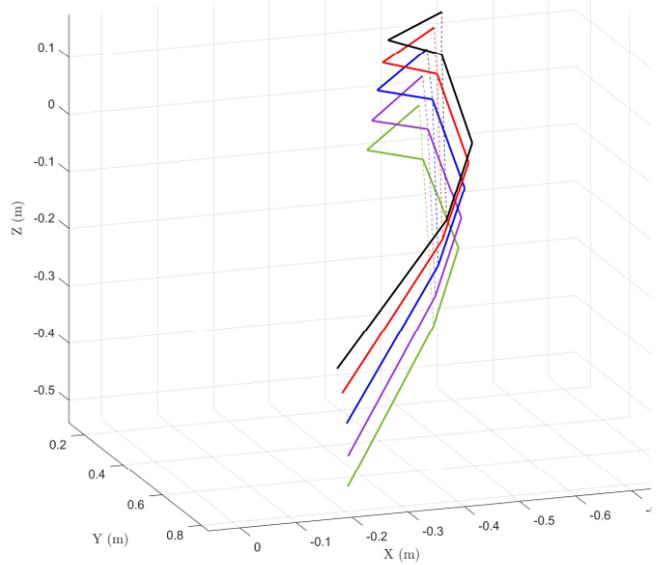

Fig. 2. Estimated poses during jump motion

The arm is composed of 3 links (including the forearm, upper-arm, and scapula), 2 joints (including the elbow and shoulder), and 4 segments and one camera lever arm (Fig. 1). Thus, the objective of the resulting estimation problem is to determine the position, velocity, and attitude of each link at each point in time (i.e. 27 variables) as well as segments and sensor biases (i.e. 33 constants), which results in 60 estimated states.

The system hardware used to evaluate the performance of the algorithms were 3 MTx Xsens modules and a smartphone camera for the SLAM system (Fig. 1b). The process noise SDs in the proposed estimation algorithms are set using the noise characteristics of the sensors presented in Table III. The initial SDs for estimation errors of position, velocity, attitude, gyroscope bias, accelerometer bias, and segments are assumed to be $10cm$, $1cm/s$, $1deg$, $0.1deg/s$, $0.1m/s^2$, and $10cm$, respectively. The SD values for position and velocity measurement noise responsible for modeling joint deformations in (20) and (25) are set at $1cm$ and $1cm/s$, respectively. The above values are established by trial and error. The SD value of the SLAM position measurement noise in (29) is set to 5cm according to the monocular ORB-SLAM3 RMSE ATE reported in [14].

To assess the effects of scapula position measurement errors and frequency disorders on algorithm performance, the EKF and SRUKF algorithms were implemented once using SLAM position output obtained from the smartphone camera data, denoted by EKF-S and SRUKF-S, and once using absolute position measurements obtained from VICON, denoted by EKF-V and SRUKF-V. The reconstructed trajectory and map point of the executed SLAM (ORB-SLAM3) as the position measurement for the EKF-S and SRUKF-S algorithms are shown in Fig. 3. The results of the proposed algorithms are shown in Fig. 4-8. The position and attitude estimation root mean square error (RMSE) for each dataset is calculated over its duration. The statistical position and attitude estimation error results for the shoulder, upper-arm, and forearm among 9 gait and 3 jump scenarios are reported in Table IV. The mean cycle run time for the EKF and SRUKF algorithms are 7.15 (*ms*) and 17.24 (*ms*), respectively, meaning that the SRUKF algorithm is about 2.4 times more computationally demanding than the EKF. Notably, the timing results were measured while the proposed algorithms were deployed under MATLAB and were run on a 3.6 GHz quad-core processor.

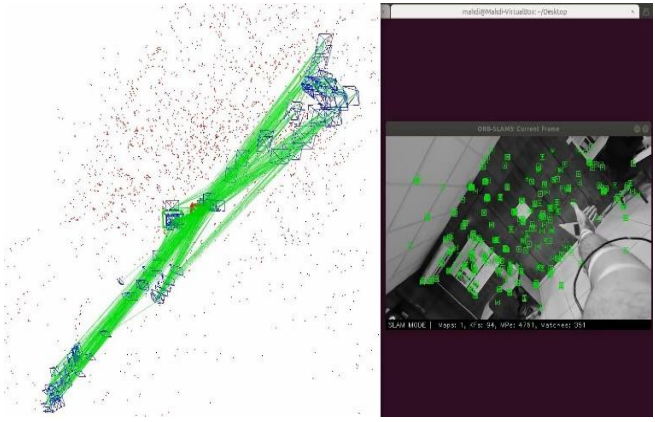

Fig. 3. ORB-SLAM3 reconstructed trajectory and map point.

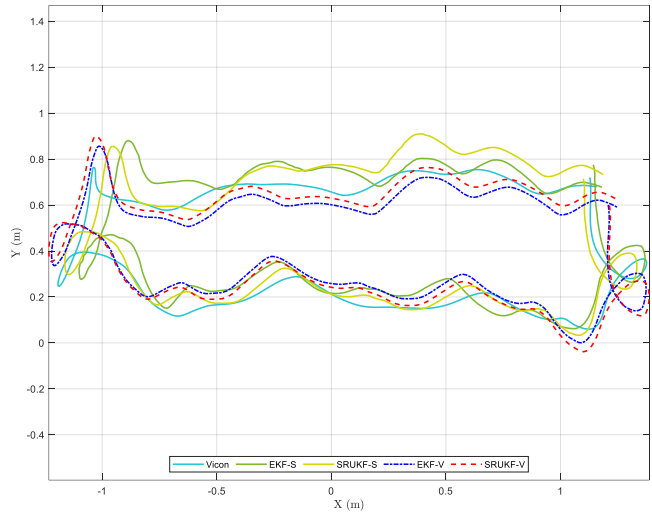

Fig. 4. Scapula link planar trajectory during jump scenario.

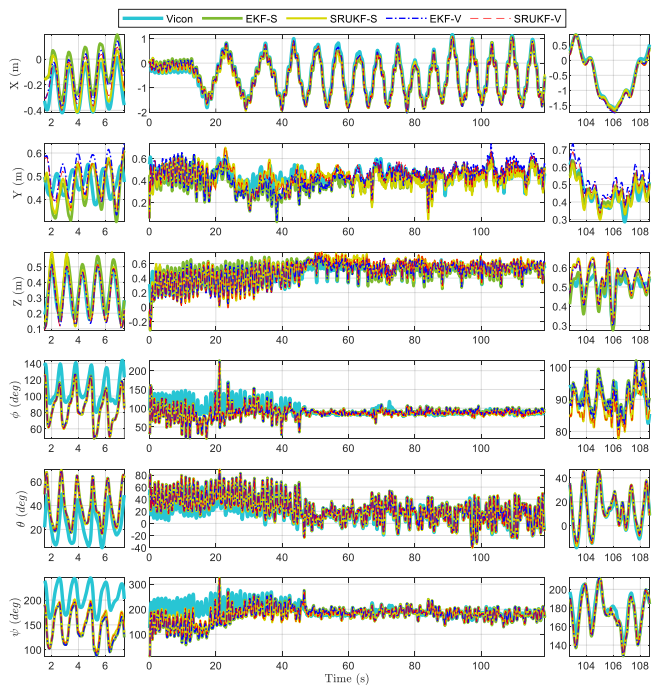

Fig. 5. Forearm link position and attitude estimation compared to the VICON during jump scenario.

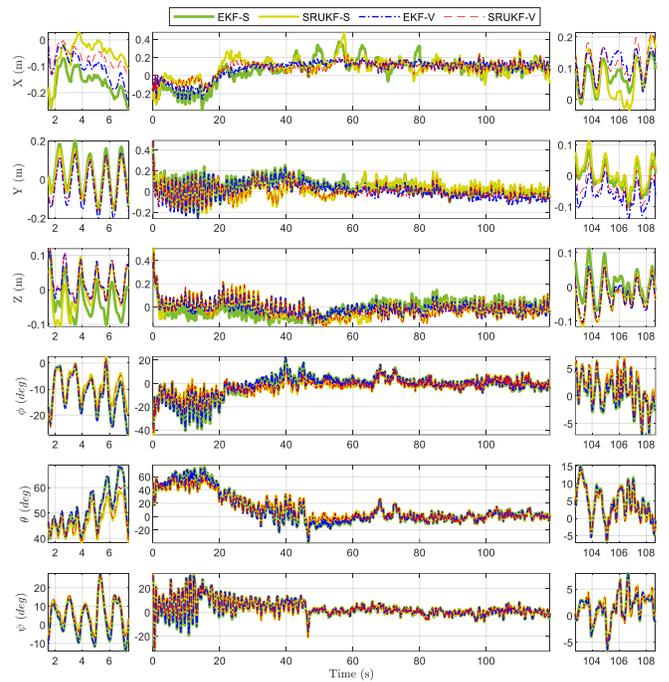

Fig. 6. Error of forearm link position and attitude estimation compared to VICON optical motion capture system during jump scenario.

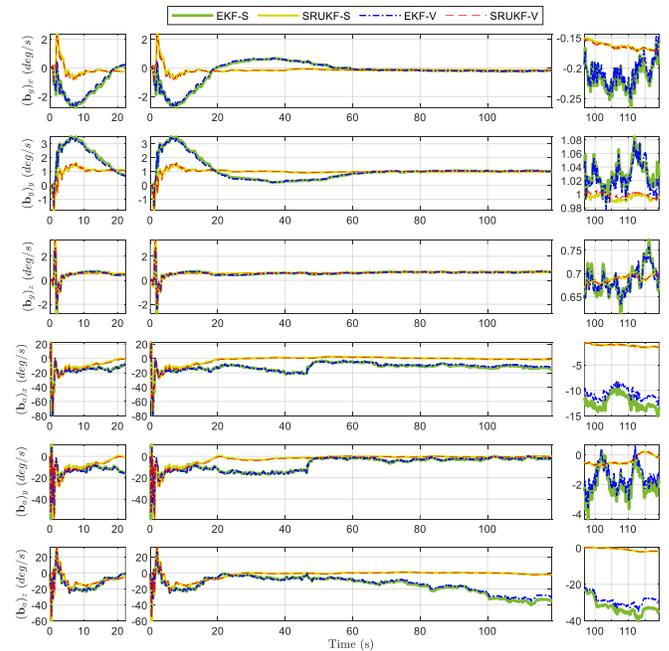

Fig. 7. Gyroscope and accelerometer bias estimation during jump scenario.

## VI. DISCUSSION

The estimated planar trajectory of the shoulder in an O-shape path gait test in comparison with the VICON is presented in Fig. 4. As expected, the EKF-V and SRUKF-V trajectories are closer to the VICON ground-truth compared to the EKF-S and SRUKF-S due to their more accurate reference position measurements. The forearm link position and attitude estimations and their estimations errors compared to the VICON are illustrated in Fig. 5-6 because it is the most challenging link due to its faster movements and distance from the camera. All the figures present position in the *n*-frame and attitude in terms of Euler angles. As shown in Fig. 6, the SLAM output error and frequency disorders caused more errors in EKF-S and SRUKF-S compared to EKF-V and SRUKF-V.

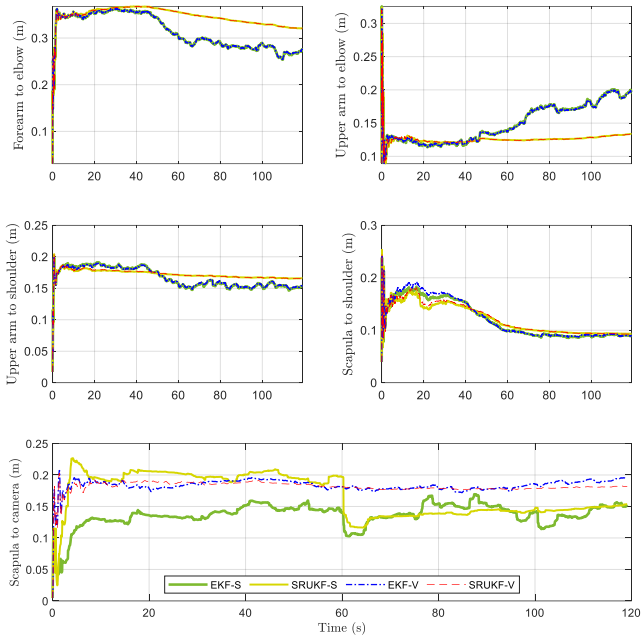

Fig. 8. Segments length estimation during jump scenario.

Gyroscope bias, accelerometer bias, and segment estimations in Fig. 7-8 show smoother convergence and higher convergence rates for SRUKF algorithms versus EKF algorithms. Moreover, according to Fig. 7, the SRUKF estimation of the biases fluctuates less than the EKF, especially in the case of accelerometer bias. Furthermore, the SRUKF estimation of sensor biases and segments is not only faster in convergence, but also more stable. There is hardly any difference in the convergence rate of the sensor biases estimation between the SLAM and VICON position measurements (Fig. 7). Assuming that no prior knowledge of the body geometry is available, initial estimations for segment lengths are set at zero. The convergence of the scapula to the camera segment length differs significantly between SLAM and VICON position measurements due to the SLAM output errors (Fig. 8). This difference decreases as the distance between the segment and the camera increases. The difference between the segment length convergence in SLAM and VICON position measurements from the shoulder joint downwards is insignificant (Fig. 8). This decrease is explicable because the biomechanical constraint measurements attenuate the effects of SLAM position measurement noise and errors.

As reported in Table IV, the minimum position and attitude estimation RMSEs across all links for the EKF-S, SRUKF-S, EKF-V, and SRUKF-V are 6.66 (*cm*) and 1.1 (*deg*), 5.87 (*cm*) and 1.11 (*deg*), 3.39 (*cm*) and 0.91 (*deg*), and 6.28 (*cm*) and 1.09 (*deg*), respectively. EKF-S is up to 80% and 40% less accurate than EKF-V, and SRUKF-S is up to 60% and 6% less accurate than SRUKF-V, in position and attitude estimation, respectively. These results were expected because of the lower estimation error and higher data frequency of the VICON reference position measurement compared to the SLAM output. It is evident that the SRUKF-S algorithm is more resistant to SLAM errors and disordered frequencies than the EKF-S.

As shown in Fig. 7-8, the convergence rate of the sensor biases and segments is faster and smoother for the SRUKF algorithms, which represents the main advantage of SRUKF algorithms in exchange for a higher computational cost. Table IV results show that the SRUKF-S is up to 17% less and 36% more accurate than the EKF-S in position and attitude estimation, respectively. In the case of EKF-V and SRUKF-V algorithms, the difference varies from link to link. In the scapula link, SRUKF-V is 8% and 6% more accurate than EKF-V, in position and attitude estimation, respectively. But in the upper-arm, link SRUKF-V is 12% and 6% less accurate than EKF-V, in position and attitude estimation, respectively.

## VII. CONCLUSION

Inertial motion capture algorithms using extended and square-root unscented Kalman filtering in combination with the SLAM method are proposed. The algorithms estimate the sensor biases and link geometries in addition to positions, velocities, and attitudes. The capability of the algorithm in estimating the link geometries and segments allows the imposition of biomechanical constraints without *a priori* knowledge of the sensors and the camera's positions. Without relying on magnetic measurements and taking advantage of SLAM position output, the system is applicable in outdoor, non-laboratory, and unknown environments. The performance of the proposed algorithms was evaluated compared to the VICON optical motion capture system during 9 fast-paced gait and 3 jump motion scenarios.

To study the effects of measurement errors and frequency variations on estimation performance, the algorithms were implemented once using the SLAM outputs and once using VICON as position measurements. Expectedly, the VICON's absolute position measurements yield less position and attitude estimation errors than SLAM, specifically in the EKF algorithms. The SRUKF algorithms present a remarkably smoother and higher convergence rate compared to the EKF. The SRUKF and EKF algorithms' estimation errors after convergence are very similar and vary from link to link. In addition, the SRUKF algorithms are 2.4 times more time-consuming than the EKF algorithms. Future research may suggest a new approach for Kalman filtering by using hybrid filters to reduce the computational cost of the SRUKF algorithm while taking advantage of its high convergence rate.

TABLE IV
STATISTICAL POSITION AND ATTITUDE ESTIMATION ERROR RESULTS

| Algorithm | Joint | Mean Position RMSE (cm) | Position RMSE SD (cm) | Min position RMSE (cm) | Max position RMSE (cm) | Mean Attitude RMSE (deg) | Attitude RMSE SD (deg) | Min attitude RMSE (deg) | Max attitude RMSE (deg) |
|---|---|---|---|---|---|---|---|---|---|
| EKF-S | Scapula | 14.03 | 2.28 | 9.87 | 17.28 | 4.54 | 1.1 | 2.53 | 6.47 |
|  | Upperarm | 12.93 | 3.21 | 6.66 | 17.42 | 2.94 | 1.48 | 1.32 | 5.73 |
|  | Forearm | 17.34 | 2.85 | 13.24 | 22.29 | 6.1 | 1.77 | 4.35 | 11.32 |
| SRUKF-S | Scapula | 14.98 | 3.97 | 5.87 | 22.55 | 4.11 | 1.11 | 2.6 | 5.87 |
|  | Upperarm | 14.65 | 3.62 | 9.51 | 21.44 | 2.2 | 1.4 | 1.05 | 6.21 |
|  | Forearm | 21.38 | 6.67 | 13.87 | 34.07 | 6.17 | 2.22 | 3.8 | 10.77 |
| EKF-V | Scapula | 10.87 | 1.52 | 9.58 | 14.38 | 4.08 | 1.12 | 2.75 | 5.91 |
|  | Upperarm | 7.1 | 1.76 | 3.39 | 9.88 | 2.07 | 0.91 | 1.25 | 3.91 |
|  | Forearm | 15.64 | 2.53 | 10.12 | 20.15 | 5.76 | 1.94 | 3.8 | 11.13 |
| SRUKF-V | Scapula | 10.15 | 1.47 | 8.33 | 12.74 | 3.9 | 1.09 | 2.4 | 5.92 |
|  | Upperarm | 9.12 | 1.8 | 6.28 | 12.45 | 2.23 | 1.52 | 1.12 | 6.56 |
|  | Forearm | 17.89 | 5.26 | 14.04 | 27.95 | 6.18 | 2.16 | 3.83 | 10.8 |